\documentclass[letterpaper]{article} 
\usepackage{aaai2026}
\usepackage{times}  
\usepackage{helvet}  
\usepackage{courier}  
\usepackage[hyphens]{url}  
\usepackage{graphicx} 
\usepackage{natbib}  
\usepackage{caption} 
\usepackage{algorithm}
\usepackage{algorithmic}
\usepackage{graphicx}
\usepackage{multirow}
\usepackage{diagbox}
\usepackage{threeparttable}
\usepackage{makecell}
\usepackage{amsmath}
\usepackage{bm}
\usepackage{booktabs}

\usepackage{times}
\usepackage{helvet}
\usepackage{courier}
\usepackage{xcolor}
\usepackage{newfloat}
\usepackage{listings}
\DeclareCaptionStyle{ruled}{labelfont=normalfont,labelsep=colon,strut=off} 
\floatstyle{ruled}
\newfloat{listing}{tb}{lst}{}
\floatname{listing}{Listing}
\makeatletter
\let\copyright@text\relax 
\makeatother

\title{IE-SRGS: An Internal-External Knowledge Fusion Framework for High-Fidelity 3D Gaussian Splatting Super-Resolution}
\author{
    Xiang Feng \textsuperscript{\rm 1,2}\thanks{Equal contribution}, Tieshi Zhong\textsuperscript{\rm 1}\footnotemark[1], Shuo Chang\textsuperscript{\rm 1}, Weiliu Wang\textsuperscript{\rm 1}, Chengkai Wang\textsuperscript{\rm 1}, Yifei Chen\textsuperscript{\rm 3}, Yuhe Wang\textsuperscript{\rm 1}, Zhenzhong Kuang\textsuperscript{\rm 1}\thanks{Corresponding author
    }, Xuefei Yin\textsuperscript{\rm 4},
 Yanming Zhu\textsuperscript{\rm 4}.
}
\affiliations{
    \textsuperscript{\rm 1}
 Hangzhou Dianzi University,
    \textsuperscript{\rm 2} ShanghaiTech University, 
    \textsuperscript{\rm 3} Tsinghua University ,
    \textsuperscript{\rm 4} Griffith University \\
    xiangfeng@hdu.edu.cn
}

\usepackage{bibentry}

\begin{document}

\makeatletter
\let\@oldmaketitle\@maketitle
\renewcommand{\@maketitle}{\@oldmaketitle
  \centering
  \includegraphics[width=0.9\linewidth]{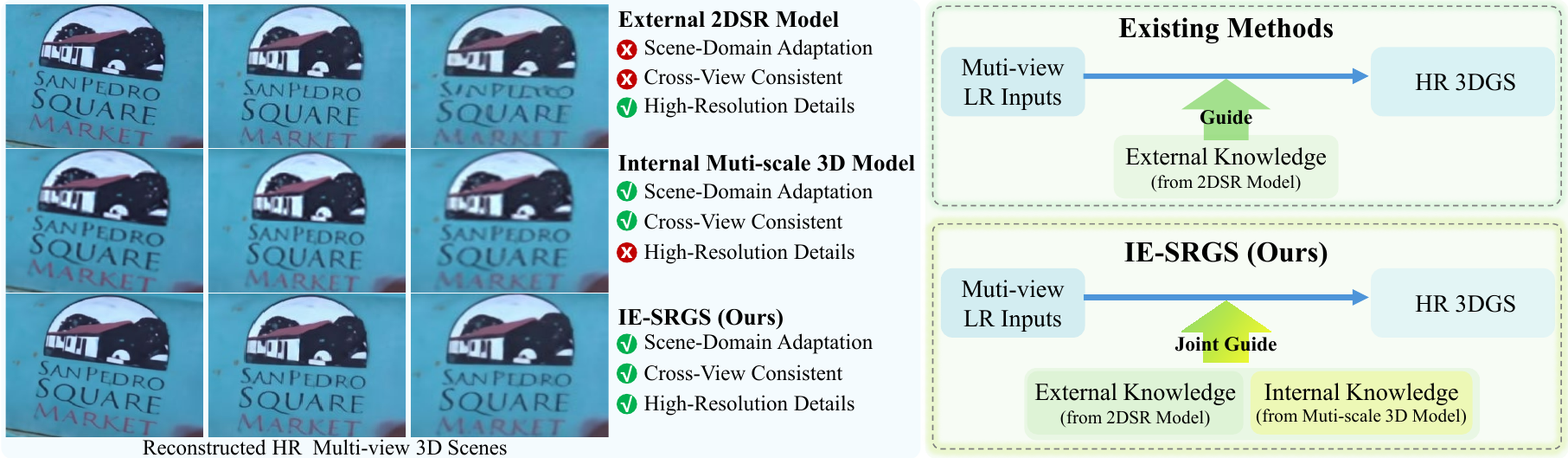}
  \captionof{figure}{Comparison between existing methods and IE-SRGS. Left: External 2D Super-Resolution (2DSR) models provide detail but lack consistency; internal 3D models offer consistency but lack details. IE-SRGS achieves consistent and detailed outputs. Right: Existing methods rely solely on external priors. IE-SRGS jointly integrates external and internal knowledge, enabling high-quality high-resolution (HR) 3D Gaussian Splatting (3DGS) from low-resolution (LR) multi-view inputs.}
  \label{fig:teaser}
  \bigskip
  }
\makeatother

\maketitle

\begin{abstract}
Reconstructing high-resolution (HR) 3D Gaussian Splatting (3DGS) models from low-resolution (LR) inputs remains challenging due to the lack of fine-grained textures and geometry. Existing methods typically rely on pre-trained 2D super-resolution (2DSR) models to enhance textures, but suffer from 3D Gaussian ambiguity arising from cross-view inconsistencies and domain gaps inherent in 2DSR models. We propose IE-SRGS, a novel 3DGS SR paradigm that addresses this issue by jointly leveraging the complementary strengths of external 2DSR priors and internal 3DGS features. Specifically, we use 2DSR and depth estimation models to generate HR images and depth maps as external knowledge, and employ multi-scale 3DGS models to produce cross-view consistent, domain-adaptive counterparts as internal knowledge. A mask-guided fusion strategy is introduced to integrate these two sources and synergistically exploit their complementary strengths, effectively guiding the 3D Gaussian optimization toward high-fidelity reconstruction. Extensive experiments on both synthetic and real-world benchmarks show that IE-SRGS consistently outperforms state-of-the-art methods in both quantitative accuracy and visual fidelity. Code will be released after review.
\end{abstract}


\section{Introduction}


3D scene reconstruction plays a crucial role in various applications. A central task in this domain is novel view synthesis (NVS), which generates photorealistic images from unseen viewpoints. 3D Gaussian Splatting (3DGS) has recently emerged as a highly efficient solution for NVS, achieving real-time, high-fidelity rendering by explicitly representing scenes as a collection of 3D Gaussians \cite{3Dgaussians}. Despite its advantages in rendering quality and speed, 3DGS struggles to reconstruct accurate scenes from low-resolution (LR) inputs due to the lack of fine-grained textures and geometric details. Moreover, acquiring, storing, and transmitting high-resolution (HR) multi-view data is often costly or infeasible in practical scenarios, motivating the need for effective super-resolution (SR) methods that enable HR 3DGS reconstruction from LR observations.

Several recent methods \cite{CROC,sequence,supergs,feng2024srgs,SuperGaussian} have explored 3DGS super-resolution, typically employing pre-trained 2D super-resolution (2DSR) models such as single-image super-resolution (SISR) or video super-resolution (VSR) to upsample LR views and generate pseudo-HR supervision for training 3DGS-based models. However, directly employing 2DSR models has two fundamental issues: (1) cross-view inconsistency, as 2D models process each view independently without enforcing multi-view consistency; and (2) domain gap, due to the discrepancy between the 2D training data and the target 3D novel scenes. Together, these issues lead to ambiguity during 3D Gaussian optimization. While prior works have attempted to address these issues through alignment, regularization, or post-processing techniques, substantial limitations remain.

In this work, we propose IE-SRGS, a novel 3DGS SR paradigm that mitigates ambiguity by jointly leveraging the complementary strength of external 2DSR priors and internal 3DGS features. Pre-trained 2DSR models offer strong HR detail priors but lack cross-view consistency and adaptability to novel 3D scenes. In contrast, multi-scale 3DGS models naturally enforce cross-view consistency and adapt to scene geometry but struggle to recover fine-grained textures from LR inputs \cite{multiscale,MipSplatting}. IE-SRGS bridges this gap through three key components: (1) applying pre-trained 2DSR and depth estimation models to generate HR images and depth maps as external knowledge; (2) employing a multi-scale 3DGS model to produce cross-view consistent, domain-adaptative counterparts as internal knowledge; and (3) introducing a mask-guided fusion strategy to effectively integrate both sources and jointly guide the 3D Gaussain optimization. By systematically fusing external and internal knowledge, IE-SRGS effectively suppresses visual artifacts, restores fine details, and enables high-quality HR 3DGS from only LR inputs. Figure \ref{fig:teaser} illustrates the paradigm shift introduced by IE-SRGS through internal-external knowledge fusion. Extensive experiments on multiple benchmark datasets demonstrate that IE-SRGS consistently outperforms state-of-the-art (SOTA) methods in both quantitative accuracy and visual quality.

Our contributions are summarized as follows:
\begin{itemize}
    \item A novel 3DGS SR paradigm, IE-SRGS, that addresses the ambiguity issue by integrating complementary internal (multi-scale 3DGS-based) and external (2DSR-based) knowledge.
    \item Added geometric priors from the depth estimation model for optimization of Gaussians in 3DGS SR.
    \item A mask-guided fusion strategy that effectively integrate internal-external guidance to supervise 3D Gaussian optimization.
    \item Extensive experimental validation on both synthetic and real-world benchmarks, demonstrating that IE-SRGS consistently outperforms SOTA methods.
\end{itemize}

\section{Related Work}
\subsection{Novel View Synthesis}
NVS aims to generate photorealistic images from unseen viewpoints using multi-view inputs \cite{lumigraph}. NeRF-based methods \cite{nerf,hollownerf,kilonerf,Mip-NeRF,bakedsdf,stylizednerf,SaNerf} achieve high visual quality, but suffer from slow rendering due to costly ray marching. Grid-based methods \cite{instant,DVGO,tensor4d,TensoRF,Zip-NeRF,NVSF} partially improve efficiency via structured discretization. Recent 3DGS methods \cite{3Dgaussians,pixelgs,deformable3dgs,shi2025mmgs,corgs,gaussianpro} have achieved real-time, high-fidelity NVS, but heavily rely on HR inputs to recover fine textures and geometry. In practice, however, acquiring HR multi-view images is costly or infeasible. To address this, methods like multi-scale reconstruction \cite{multiscale} and Mip-Splatting \cite{MipSplatting} have been proposed to improve detail handling, but still struggle to recover high-frequency content without external priors.

\subsection{Single-Image and Video Super-Resolution}
SISR and VSR are widely used to generate external HR priors for 3DGS SR. SISR enhances spatial details from LR images using CNNs \cite{EDSR,rcan,dingmodal}, Transformers \cite{Swinir}, or GAN/diffusion models \cite{SRGAN,ESRGAN,lg-diff,SR3}. VSR extends this by using temporal information across frames through motion-aware architectures \cite{basicvsr,basicvsr++,vrt}. While effective in 2D tasks, these models face challenges when applied to multi-view 3D reconstruction. They process each view independently, introducing cross-view inconsistencies, and often suffer from performance degradation due to domain gaps between training data and target 3D scenes. These limitations motivate our framework, which explicitly addresses cross-view consistency and domain adaptation for 3DGS SR.

\begin{figure*}
    \centering
    \includegraphics[width=0.91\textwidth]{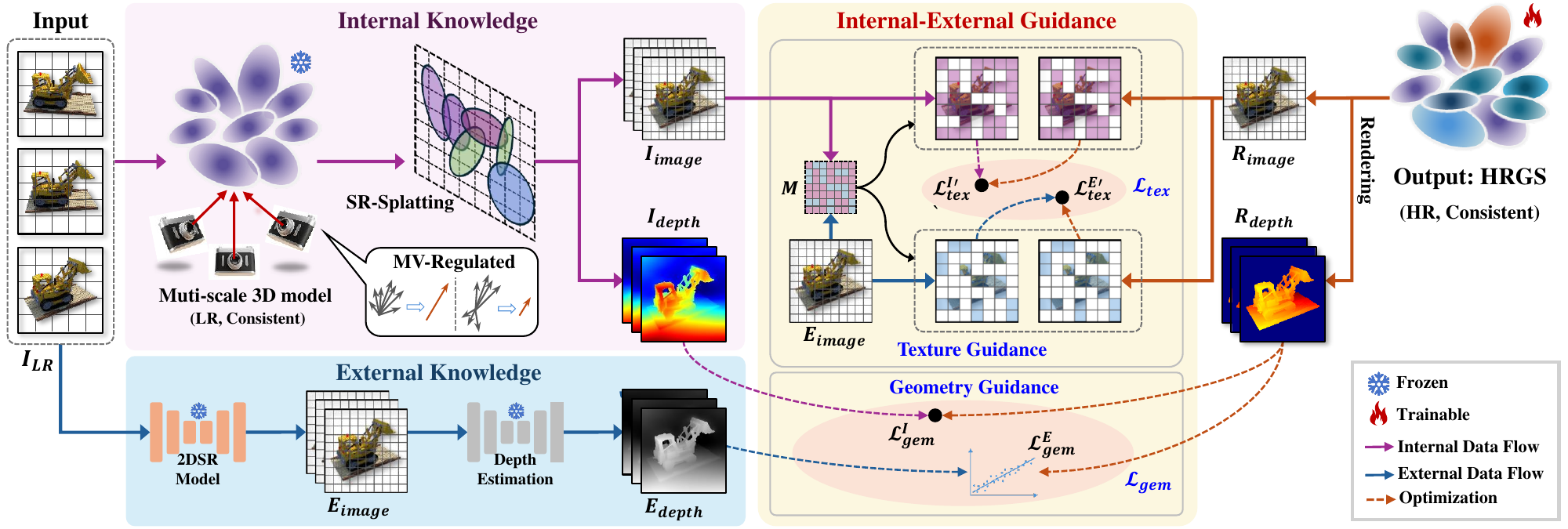}
    \caption{Overall framework of the proposed IE-SRGS. Given LR multi-view images, IE-SRGS extracts external guidance from pre-trained 2DSR models and internal guidance from a multi-scale 3D model. A mask-guided internal-external guidance integration provides structured, adaptive supervision for final HR 3DGS optimization.}
    \label{fig:framework}
\end{figure*}

\subsection{3D Super-Resolution}
High-quality 3D reconstruction from LR inputs remains challenging for both NeRF- and 3DGS-based methods. Earlier NeRF-based works \cite{NeRF-SR,zs-srt,CROC,disr} leverage super-sampling or 2DSR priors to improve texture quality, but often suffer from inefficiency or limited adaptability. Recent 3DGS-based methods \cite{SuperGaussian,sequence,supergs,feng2024srgs,gaussiansr} enable real-time rendering without extensive scene-specific fine-tuning, such as SRGS \cite{feng2024srgs} and GaussianSR \cite{gaussiansr} methods. However, they rely mainly on external priors, largely ignoring internal 3D consistency and scene-specific adaptation. To address these limitations, we propose IE-SRGS, which integrates external high-frequency priors with internal 3D consistency from multi-scale 3DGS representations, and explicitly incorporates geometric depth priors to further enhance reconstruction quality.

\section{Preliminaries}
\noindent\textbf{3D Gaussian Splatting.} 
3DGS \cite{3Dgaussians} represents a 3D scene using a set of anisotropic 3D Gaussians $\mathcal{G} = \{\mathbf{g}_1, \mathbf{g}_2,..., \mathbf{g}_{N} \}$, where each Gaussian is parameterized by a 3D position $\bm{\mu}$, covariance matrix $\bm{\Sigma}$, color $\bm{c}$, and opacity $\alpha$. Each Gaussian defines a density distribution in space:
\begin{equation}
\mathbf{g}^\text{3D}(\bm{x}) 
= \exp\left(
-\frac{1}{2}(\bm{x}-\bm{\mu})^\top\bm{\Sigma}^{-1}(\bm{x}-\bm{\mu})
\right),
\label{eq:gaussian_3d}
\end{equation}
where $\bm{x}$ denotes a 3D location. These Gaussians are projected into screen space and rasterized using splatted alpha blending to render novel views. In this work, we aim to enhance HR 3DGS reconstruction from LR inputs by improving the texture and geometric quality of Gaussian primitives via internal–external knowledge fusion.


\noindent\textbf{Mip-Splatting.}
Mip-Splatting \cite{MipSplatting} enhances 3DGS rendering stability by applying a 3D smoothing operation that suppresses aliasing and high-frequency noise under magnification. Specifically, each 3D Gaussian is convolved with a low-pass filter before projection:
\begin{equation}
\mathbf{g}^{\text{3D}}_{\text{reg}}(\bm{x}) = (\mathbf{g}^{\text{3D}} \otimes \mathbf{g}_{\text{low}})(\bm{x}),
\end{equation}
which yields a regularized Gaussian with enlarged covariance, effectively controlling the sampling rate and smoothing sharp artifacts. This operation preserves geometric consistency across views and improves robustness to domain variations. In our framework, we leverage this multi-scale rendering consistency to mitigate cross-view ambiguity.


\section{Method}
\noindent\textbf{Overview.} The overall framework of IE-SRGS is shown in Figure \ref{fig:framework}. Given a set of LR multi-view images, we first apply a pre-trained 2DSR model and a depth estimator to generate HR images and depth maps, forming the external guidance rich in texture details. Meanwhile, a multi-scale 3DGS model is constructed directly from LR inputs and optimized via multi-view regularization to ensure cross-view consistency and scene adaptation. We then apply a SR splatting to generate internal, upscaled images and depth maps, which serve as the internal guidance. Finally, we introduce a mask-guided fusion strategy to integrate internal and external knowledge, leveraging their complementary strengths in texture and geometry to guide HR 3DGS optimization.


\subsection{External Knowledge for HR Detail Restoration}
\label{Exter}
Reconstructing high-quality HR 3DGS from LR inputs is challenging due to lacking HR fine textures and accurate geometry. To address this, we propose to leverage external knowledge from pre-trained 2DSR and depth estimation models, which provide strong priors learned from large-scale datasets. Specifically, we use SwinIR \cite{Swinir} to generate super-resolved images with rich texture details, and Depth Anything V2 \cite{depth_anything} to estimate depth maps from these images, offering geometric cues. The resulting super-resolved image and depth map are denoted as $E_{\text{image}}$ and $E_{\text{depth}}$, respectively, and serve as external guidance for optimizing the 3D Gaussians.

To inject external knowledge into the 3DGS model, we supervise the optimization of its Gaussian using both the texture and geometric information provided by $E_{\text{image}}$ and $E_{\text{depth}}$. For texture guidance, we compare the rendered image $R_{\text{image}}$, obtained by projecting the current Gaussian onto the target view during optimization, with the external reference $E_{\text{image}}$ using a weighted sum of $L_{1}$ loss and D-SSIM losses \cite{3Dgaussians} $\mathcal{L}_{\mathrm{ds}}$:
\begin{equation}
    \mathcal{L}^E_{\text{tex}}=(1-\lambda)\mathcal{L}_{1}({E}_{\text{image}}, {R}_{\text{image}})+\lambda\mathcal{L}_{\text{ds}}({E}_{\text{image}}, {R}_{\text{image}}),
\end{equation}
where $\lambda$ balances pixel-wise accuracy and structural similarity. For geometric supervision, we align the rendered depth $R_{\text{depth}}$, obtained via splatted rasterization during Gaussian optimization, with $E_{\text{depth}}$ using a relaxed relative loss \cite{sparsegs} based on Pearson correlation:
\begin{equation}
\mathcal{L}^E_{\text{gem}}=\frac{1}{N}\sum_{i=1}^{N}(1-\frac{\mathrm{Cov}({{{R}_{\text{depth}}^{i}}},{{E}}_{\text{depth}}^{i})}{\sqrt{\mathrm{Var}({{{R}_{\text{depth}}^{i}}})\operatorname{Var}({{E}}_{\text{depth}}^{i})}}),
\end{equation}
where $N$ is the number of pixels, and $\mathrm{Cov}(\cdot)$ and $\mathrm{Var}(\cdot)$ denote covariance and variance, respectively. Together, these losses guide 3D Gaussians to capture fine textures and accurate geometry from external knowledge.

 \begin{figure*}
    \centering
    \includegraphics[width=0.9\textwidth]{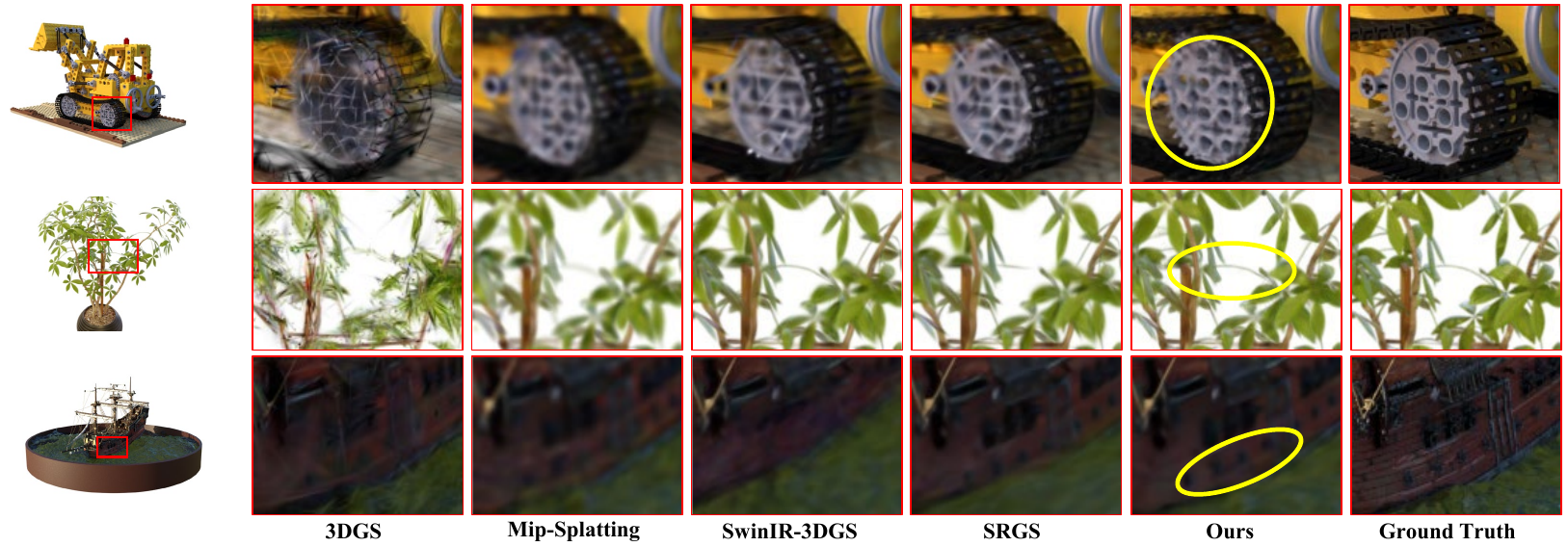}
    \caption{Qualitative comparisons of 4$\times$ 3D super-resolution on the NeRF Synthetic dataset. IE-SRGS achieves sharper textures and higher fidelity compared to SOTA methods 3DGS, Mip-Splatting, SwinIR-3DGS, and SRGS.}
    \label{fig:blender}
\end{figure*}

\subsection{Internal Knowledge for Ambiguity Correction}
\label{Inter}
While external knowledge offers rich texture and geometry priors for Gaussian optimization, it suffers from: (1) cross-view inconsistency, as 2D models process each view independently without enforcing 3D coherence; and (2) domain gaps between training data and target scenes. These limitations introduce ambiguity in Gaussian optimization. To address this, we introduce internal knowledge from a multi-scale 3DSR model that inherently enforces cross-view consistency and adaptability. We adopt Mip-Splatting \cite{MipSplatting} as the internal backbone for its anti-aliasing capability and scale-consistent representation.

To extract internal guidance, we first train a multi-scale 3DGS model using LR multi-view inputs. To further enhance cross-view consistency, we adopt Multi-View Regulation (MV-Regulation) \cite{MVGS}, which supervises multiple views jointly during optimization. This strategy reduces overfitting to individual views and promotes geometric coherence across viewpoints. We then apply SR-Splatting to generate HR internal references. Specifically, 3D Gaussians are projected onto a 2D screen-space plane and upsampled using a predefined scaling factor. The upsampled splats are rasterized to produce internally scaled images and depth maps, denoted as ${I}_{\text{image}}$ and ${I}_{\text{depth}}$, which serve as internal guidance for Gaussian optimization.
To correct the ambiguities introduced by external supervision, we supervise the rendered image ${R}_{\text{image}}$ and depth map ${R}_{\text{depth}}$ using the internal references ${I}_{\text{image}}$ and ${I}_{\text{depth}}$ through  dedicated internal texture and geometry losses:
\begin{equation}
    \mathcal{L}^I_{\text{tex}}=(1-\lambda)\mathcal{L}_{1}({I}_{\text{image}}, {R}_{\text{image}})+\lambda\mathcal{L}_{\mathrm{ds}}({I}_{\text{image}}, {R}_{\text{image}}),
\end{equation}
\begin{equation}
    \mathcal{L}^I_{\text{gem}}=\mathcal{L}_{1}({I}_{\text{depth}}, {R}_{\text{depth}}).
\end{equation}
These losses guide the 3D Gaussians toward consistent and scene-adaptive reconstructions, complementing the HR details provided by external knowledge.

\begin{figure*}
    \centering
    \includegraphics[width=0.9\textwidth]{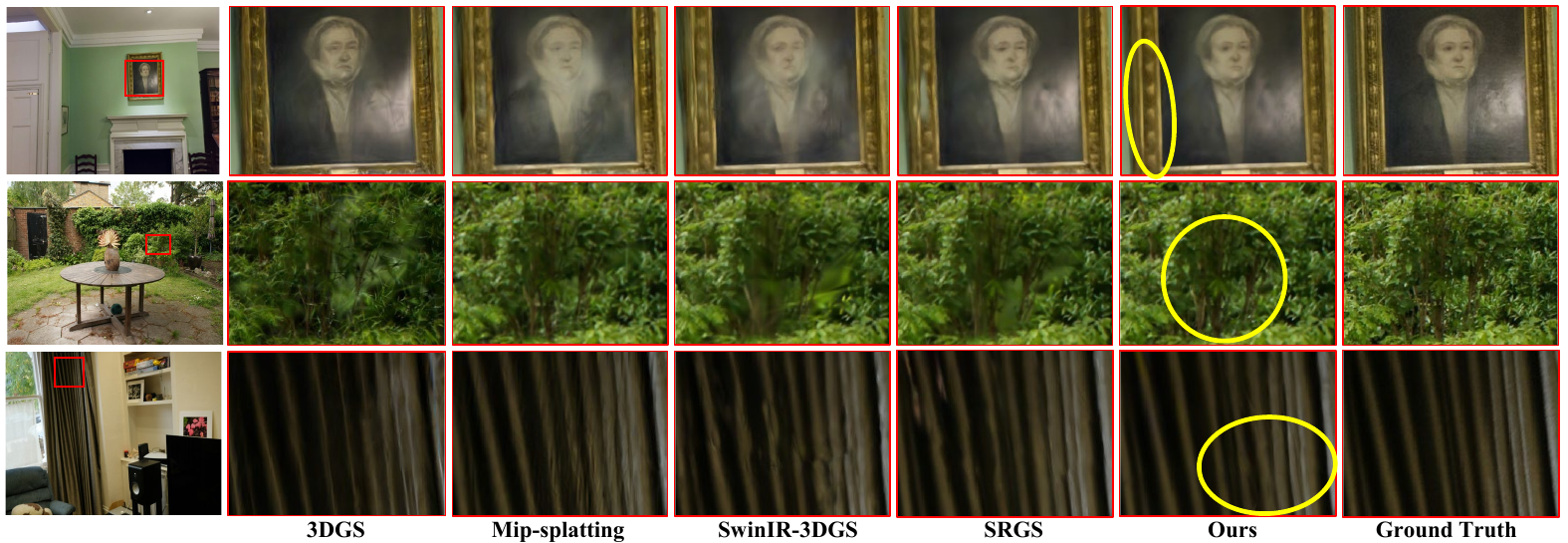}
    \caption{Qualitative comparisons of 4$\times$ 3D super-resolution on real-world datasets. IE-SRGS reconstructs finer textures and preserves more structural details compared to SOTA methods 3DGS, Mip-Splatting, SwinIR-3DGS, and SRGS.}
    \label{fig:real-world}
\end{figure*}

\subsection{Internal–External Fusion for HRGS Optimization}
\label{Gration}
To fully exploit the complementary strengths of internal and external guidance, we propose integrating them through a unified supervision framework. This fusion enables accurate texture restoration and robust geometric reconstruction, and is structured into two components: Texture Guidance Integration and Geometric Guidance Integration. 

\noindent\textbf{Texture Guidance Integration.} 
Considering that the inconsistencies and local artifacts introduced by 2DSR models are typically localized, and internal guidance ensures cross-view consistency but lacks HR details, we propose a mask-guided integration strategy for texture supervision. This can effectively address inconsistencies while preserving the high-quality texture details. Specifically, an uncertainty map $D(p)$ is computed at each pixel $p$ to measure the discrepancy between $I_{\text{image}}$ and $E_{\text{image}}$:
\begin{equation}
    D(p)=\frac{|{I}_{\text{image}}(p)-E_{\text{image}}(p)|}{I_{\text{image}}(p)+\epsilon},
\end{equation}
where $\epsilon$ is a small constant ($10^{-6}$) for numerical stability. A binary mask $M(p)$ is then computed by thresholding $D(p)$:
\begin{equation}
    M(p) =
    \begin{cases}
    1, &  D(p) \geq T \\
    0, &  D(p) < T
    \end{cases}\quad.
\end{equation}
This mask guides the texture supervision process: regions with high discrepancy are trained using internal references, while the the remaining areas follow external guidance. The final texture loss is defined as:
\begin{equation}
 \mathcal{L}_{\text{tex}} = \mathcal{L}^{I'}_{\text{tex}}+\mathcal{L}^{E'}_{\text{tex}},
\end{equation}
where $\mathcal{L}^{I'}_{\text{tex}}=\mathcal{L}^I_{\text{tex}} \odot M(p)$ and $ \mathcal{L}^{E'}_{\text{tex}}=\mathcal{L}^E_{\text{tex}} \odot (1 - M(p))$.


\noindent\textbf{Geometric Guidance Integration.} As geometric structures are relatively coarse and less sensitive to local variance, we directly combine internal and external geometric losses via a weighted sum, controlled by balancing factors $\lambda_i$ and $\lambda_e$:
\begin{equation}
 \mathcal{L}_{\text{gem}} = \lambda_{i} \mathcal{L}^I_{\text{gem}}  + \lambda_{e}\mathcal{L}^E_{\text{gem}},
\end{equation}

The final loss for HR 3DGS optimization is defined as:
\begin{equation}
 \mathcal{L}_{\text{final}} =  \mathcal{L}_{\text{mask}}  + \mathcal{L}_{\text{gem}} .
\end{equation}
Through joint internal-external guidance at both texture and geometry levels, IE-SRGS effectively resolves ambiguities, preserves HR details, and achieves high-fidelity 3D reconstruction from LR inputs.

\section{Experiments}
\subsection{Datasets and Evaluation Metrics}
We evaluate IE-SRGS on 21 scenes from four public datasets to demonstrate its robustness and generalization. Specifically, we use 9 real-world indoor and outdoor scenes from Mip-NeRF 360 \cite{mipnerf360}, 2 from Deep Blending \cite{DeepBlending}, 2 from Tanks\&Temples \cite{tanks}, and 8 synthetic scenes from NeRF Synthetic \cite{nerf}. To construct LR inputs, training views are downsampled using bicubic interpolation, by a factor of 8 for Mip-NeRF 360 (for $4\times$ 3D super-resolution) and by a factor of 4 for the other datasets, following SRGS \cite{feng2024srgs} for fair comparison. We assess reconstruction quality using Peak Signal-to-Noise Ratio (PSNR), Structural Similarity Index (SSIM) \cite{SSIM}, and Learned Perceptual Image Patch Similarity (LPIPS) \cite{lpip}.

\subsection{Implementation Details}
IE-SRGS is built upon the open-source Mip-Splatting codebase, with modification to the Gaussian rasterization module for depth map rendering. For internal model training, we use 3 randomly sampled views in MV-Regulation and optimize the multi-scale 3DGS model for 30,000 iterations using the same hyperparameters as Mip-Splatting \cite{MipSplatting}. We set $\lambda_{i} =0.001$ and $\lambda_{e}=0.0001$ for internal and external supervision and use a discrepancy threshold $T$ of 0.9 for real-world scenes and 0.6 for synthetic scenes. All experiments are run on a single NVIDIA RTX 4090 GPU.

\subsection{Performance Comparison}
We compare IE-SRGS with a range of SOTA 3D SR methods, including NeRF-SR \cite{NeRF-SR}, CROC \cite{CROC}, DiSR-NeRF \cite{disr}, FastSR-NeRF \cite{fastsr}, GaussianSR \cite{gaussiansr}, SuperGaussian \cite{SuperGaussian}, SRGS \cite{feng2024srgs}, and Sequence Matters \cite{sequence}. SRGS is the most relevant baseline and Sequence Matters reports the best performance. For methods without public code (CROC, FastSR-NeRF, GaussianSR, SuperGaussian), we cite results from their papers. For real-world datasets lacking their results, we include comparisons with 3D Gaussian Splatting (3DGS) \cite{3Dgaussians}, Mip-Splatting \cite{MipSplatting}, and a baseline variant SwinIR-3DGS.

\begin{table}[htbp]
    \centering
    \begin{threeparttable}
        \small 
        \setlength{\tabcolsep}{2pt}
        \renewcommand{\arraystretch}{0.6}
        \begin{tabular}{lccc}
            \toprule[0.8pt]
            \textbf{Methods} & \multicolumn{3}{c}{\textbf{NeRF Synthetic}} \\
            \cline{2-4}
            & PSNR$\uparrow$ & SSIM$\uparrow$ & LPIPS$\downarrow$ \\ 
            \midrule[0.8pt]
            3DGS \cite{3Dgaussians} & 21.77 & 0.867 & 0.104 \\
            SwinIR-3DGS \cite{Swinir} & 30.38 & 0.945 & 0.059 \\
            NeRF-SR \cite{NeRF-SR} & 28.90 & 0.927 & 0.099 \\
            DiSR-NeRF \cite{disr} & 26.00 & 0.890 & 0.123 \\
            FastSR-NeRF \cite{fastsr} & 30.47 & 0.944 & 0.075 \\
            CROC \cite{CROC} & 30.71 & 0.945 & 0.067 \\
            Mip-splatting \cite{MipSplatting} & 24.59 & 0.909 & 0.101 \\
            GaussianSR \cite{gaussiansr} & 28.37 & 0.924 & 0.087 \\
            SuperGaussian \cite{SuperGaussian} & 28.44 & 0.923 & 0.067 \\
            SRGS \cite{feng2024srgs} & 30.83 & 0.948 & 0.056 \\
            Ours & \textbf{30.97} & \textbf{0.952} & \textbf{0.054} \\
            \midrule
            Upper Bound & 33.37 & 0.969 & 0.032 \\
            \bottomrule[0.8pt]
        \end{tabular} 
        \begin{tablenotes}
            \footnotesize
            \item Note: Upper Bound refers to the results obtained by training HR 3DGS directly with HR inputs.
        \end{tablenotes}
                \caption{Quantitative comparison of 4$\times$ 3D super-resolution results on NeRF Synthetic dataset.}
        \label{tab:blender}
    \end{threeparttable}
\end{table}

\begin{table*}[htbp]
    \centering
    \begin{threeparttable}
        \small 
        \setlength{\tabcolsep}{4pt} 
        \renewcommand{\arraystretch}{0.6}
        \begin{tabular}{@{}l *{3}{cccc} @{}}
            \toprule[0.8pt]
            \textbf{Methods} & \multicolumn{3}{c}{\textbf{Mip-NeRF360}} &
                           & \multicolumn{3}{c}{\textbf{Deep Blending}} &
                           & \multicolumn{3}{c}{\textbf{Tanks\&Temples}} &\\
            \cline{2-4}\cline{6-8}\cline{10-12}
            & PSNR$\uparrow$ & SSIM$\uparrow$ & LPIPS$\downarrow$ &
            & PSNR$\uparrow$ & SSIM$\uparrow$ & LPIPS$\downarrow$ &
            & PSNR$\uparrow$ & SSIM$\uparrow$ & LPIPS$\downarrow$ &\\
            \midrule[0.8pt]
            3DGS \cite{3Dgaussians} 
            & 20.72 & 0.617 & 0.396&
            & 27.02 & 0.851 & 0.304&
            & 19.62 & 0.715 & 0.337& \\
            
            SwinIR-3DGS \cite{Swinir}
            & 26.68 & 0.762 & 0.299&
            & 29.29 & 0.892 & 0.279&
            & 23.32 & 0.805 & 0.281& \\
            
            Mip-splatting \cite{MipSplatting}
            & 26.43 & 0.754 & 0.304&
            & 28.93 & 0.885 & 0.283&
            & 23.04 & 0.790 & 0.293& \\
            
            GaussianSR \cite{gaussiansr}
            & 25.60 & 0.663 & 0.368&
            & 28.28 & 0.873 & 0.307&
            & -- & -- & -- &\\
            
            SRGS \cite{feng2024srgs}
            & 26.88 & 0.767 & 0.286&
            & 29.49 & 0.896 & 0.275&
            & 23.41 & 0.807 & 0.278& \\

            Squence Matters \cite{sequence}
            & 27.02 & 0.774 & 0.279&
             & -- & -- & --&
            & 23.43 & 0.808 & 0.274&
            \\
            
            Ours
            & \textbf{27.15} & \textbf{0.779} & \textbf{0.278}&
            & \textbf{29.63} & \textbf{0.899} & \textbf{0.271}&
            & \textbf{23.52} & \textbf{0.810} & \textbf{0.274} &\\
            \midrule
            Upper Bound
            & 27.23 & 0.797 & 0.254&
            & 29.73 & 0.905 & 0.243&
            & 23.51 & 0.828 & 0.242 &\\
            \bottomrule[0.8pt]
        \end{tabular}
        \begin{tablenotes}
            \footnotesize
            \item Note: Upper Bound refers to the results obtained by training HR 3DGS directly with HR inputs.
        \end{tablenotes}
    \end{threeparttable}
            \caption{Quantitative comparison of 4$\times$ 3D super-resolution results on three real-world datasets Mip-NeRF360, Deep Blending, and Tanks\&Temples evaluated by PSNR, SSIM, and LPIPS.}
        \label{tab:real_world}
\end{table*}

\begin{table}[t]
    \centering
    \small 
    \setlength{\tabcolsep}{2pt} 
    {
    \renewcommand{\arraystretch}{0.6}
    \begin{tabular}{@{}lccc@{}}
        \toprule[0.8pt]
        \textbf{Methods} & \multicolumn{3}{c}{\textbf{MipNeRF-360}} \\
        \cline{2-4}
        & PSNR$\uparrow$ & SSIM$\uparrow$ & LPIPS$\downarrow$ \\
        \midrule[0.8pt]
        Mip-Splatting (Baseline) & 26.43 & 0.754 & 0.304 \\
        \addlinespace[0.1em]
        $+$ MV-Regulation & 26.68 & 0.757 & 0.297 \\
        \midrule
        Mip-Splatting (Baseline) & 26.43 & 0.754 & 0.304 \\
        \addlinespace[0.1em]
        $+$ External Texture Guidance ($E_\text{image}$) & 26.69 & 0.762 & 0.300 \\
        \addlinespace[0.1em]
        $+$ External Geometric Guidance ($E_\text{depth}$) & 26.72 & 0.763 & 0.299 \\
        \addlinespace[0.1em]
        $+$ Internal Texture Guidance ($I_\text{image}$) & 27.00 & 0.775 & 0.283 \\
        \addlinespace[0.1em]
        $+$ Internal Geometric Guidance ($I_\text{depth}$) & 27.05 & 0.775 & 0.282 \\
        \addlinespace[0.1em]
        $+$ Mask-Guided Texture Integration & \textbf{27.15} & \textbf{0.779} & \textbf{0.278} \\
        \bottomrule[0.8pt]
    \end{tabular} }
        \caption{Ablation study on component effectiveness of IE-SRGS. MV-Regulation is applied to Mip-Splatting to form a multi-view consistency baseline, while guidance components are progressively added without MV-Regulation.}
    \label{tab:ablation}
\end{table}

\subsubsection{Quantitative Comparison}
Table \ref{tab:blender} presents results on the NeRF Synthetic dataset for $4\times$ 3D SR, a most common benchmark for SOTA methods. As shown, IE-SRGS outperforms all SOTA methods across all metrics, achieving the best performance. Notably, IE-SRGS improves upon its backbone Mip-Splatting by 25.9\% in PSNR, 4.73\% in SSIM, and 46.5\% in LPIPS, clearly demonstrating the effectiveness of internal-external knowledge fusion. Moreover, its performance closely approaches the Upper Bound trained on HR inputs, highlighting its strong capability to recover accurate textures and geometries from LR inputs.

Table \ref{tab:real_world} reports results on three real-world datasets: Mip-NeRF360, Deep Blending, and Tanks\&Temples. While many SOTA methods avoid evaluation on real-world scenes due to increased complexity and domain shifts, we evaluate across all datasets to thoroughly assess the robustness of our method. As shown, IE-SRGS again achieves the best performance across all metrics and datasets, closely approaching the Upper Bound. These results highlight its strong generalization under real-world LR constraints and validate our core claim that joint internal–external guidance enables consistent, high-quality 3D SR across diverse domains.

\subsubsection{Qualitative Comparison}
Figure \ref{fig:blender} and Figure \ref{fig:real-world} show qualitative results for $4\times$ 3D SR on synthetic and real-world datasets, respectively. As observed, standard 3DGS exhibits severe blurring due to the absence of HR details. Mip-Splatting improves consistency but still lacks fine textures, revealing the limitations of internal knowledge alone. Methods using external knowledge, such as SwinIR-3DGS and SRGS, recover partial details but introduce noticeable artifacts and distortions, especially in regions with complex structures. In contrast, IE-SRGS consistently generates sharper textures, more accurate geometry, and visually coherent results across all datasets, validating the effectiveness of internal–external fusion. Reconstructed depth maps are compared in the Supplementary.

\begin{figure*}
    \centering
    \includegraphics[width=0.8\linewidth]{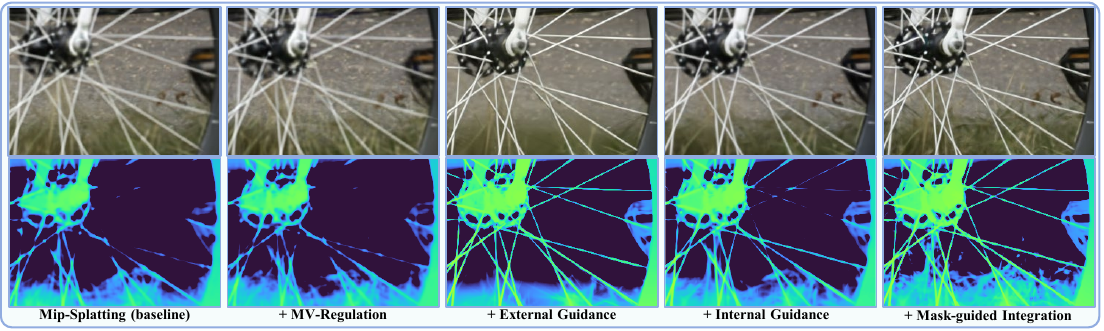}
    \caption{Qualitative results of the component effectiveness analysis. MV-Regulation is first added to Mip-Splatting to form a consistency baseline, while external, internal, and joint guidance components are progressively added without MV-Regulation.}
    \label{fig:ablation}
\end{figure*}

\subsection{Ablation Study}
\subsubsection{Component Contribution Analysis}
To evaluate the effectiveness of each component in IE-SRGS, we conduct ablation studies on the Mip-NeRF360 dataset. We first enhance Mip-Splatting with MV-Regulation to serve as a reference baseline for evaluating the impact of enforcing multi-view consistency. Then, starting from the Mip-Splatting baseline, we progressively introduce external texture guidance ($E_\text{image}$), external geometric guidance ($E_\text{depth}$), internal texture guidance ($I_\text{image}$), internal geometric guidance ($I_\text{depth}$), and finally, mask-guided texture integration.

\begin{table}[t]
    \centering
    \small 
    \setlength{\tabcolsep}{2pt} 
    {
    \renewcommand{\arraystretch}{0.6}
    \begin{tabular}{llccc}
        \toprule[0.8pt]
        & \multirow{2}{*}{\textbf{Backbones}} & \multicolumn{3}{c}{\textbf{MipNeRF-360}} \\
        \cline{3-5}
        &  & PSNR $\uparrow$ & SSIM $\uparrow$ & LPIPS $\downarrow$ \\
        \midrule[0.8pt]
        \multirow{4}{*}{\rotatebox{90}{External}} & PSRT \cite{rethinkingvsr} & 25.19 & 0.697 & 0.353 \\
        & Ours (PSRT) & \textbf{25.67} & \textbf{0.725} & \textbf{0.310} \\
        & SwinIR \cite{Swinir} & 25.23 & 0.699 & 0.332 \\
        & Ours (SwinIR) & \textbf{25.73} & \textbf{0.729} & \textbf{0.306} \\
        \midrule
        \multirow{4}{*}{\rotatebox{90}{Internal}} & Analy-Splatting \cite{analytic} & 24.14 & 0.629 & 0.411 \\
        & Ours (Analy-Splatting) & \textbf{25.48} & \textbf{0.716} & \textbf{0.324} \\
        & Mip-Splatting \cite{MipSplatting} & 25.04 & 0.687 & 0.349 \\
        & Ours (Mip-Splatting) & \textbf{25.73} & \textbf{0.729} & \textbf{0.306} \\
        \bottomrule[0.8pt]
    \end{tabular} }
        \caption{Performance comparison across different backbones with and without our joint internal-external guidance.}
    \label{tab:backbones}
\end{table}

\begin{table*}[t]
    \centering
    \small 
    \setlength{\tabcolsep}{3.5pt} 
    {
    \renewcommand{\arraystretch}{0.5}
    \begin{tabular}{llccccccc}
    \toprule[0.8pt]
    \textbf{Datasets} & \textbf{Methods} & \textbf{External Train} & \textbf{Internal Train} & \textbf{Depth Estimation} & \textbf{HRGS Train} & \textbf{Total Train Time} & \textbf{Inference (FPS)} \\
    \toprule[0.8pt]
    \multirow{2}{*}{NeRF Synthetic} & SRGS & 45.0s & - & - & 13min11s & 13mins 56s & 191 \\
    & IE-SRGS & 45.0s & 6min40s & 23.9s & 11mins42s & 19mins30s & \textbf{260} \\
    \midrule
    \multirow{2}{*}{MipNeRF-360} & SRGS & 2min39s & - & - & 43min12s & 45mins 51s & 92 \\
    &IE-SRGS & 2min39s & 10min30s & 48.3s & 40mins20s & 54mins17s & \textbf{119} \\
    \bottomrule[0.8pt]
    \end{tabular} }
        \caption{Time breakdown and inference speed comparison of SRGS vs. IE-SRGS on NeRF Synthetic and MipNeRF-360.}
    \label{tab:runtime}
\end{table*}

\begin{table}[t]
    \centering
    \small
    \setlength{\tabcolsep}{4pt} 
    {
    \renewcommand{\arraystretch}{0.6}
    \begin{tabular}{@{}lccc@{}}
        \toprule[0.8pt]
        \textbf{Methods} & \multicolumn{3}{c}{\textbf{MipNeRF-360}} \\
        \cline{2-4}
        & PSNR$\uparrow$ & SSIM$\uparrow$ & LPIPS$\downarrow$ \\
        \midrule[0.8pt]
        Mip-Splatting \cite{MipSplatting} & 25.02 & 0.728 & 0.417 \\
        SRGS \cite{feng2024srgs} & 25.27 & 0.741 & 0.405 \\
        Ours & \textbf{25.64} & \textbf{0.755} & \textbf{0.386} \\
        \bottomrule[0.8pt]
    \end{tabular}}
        \caption{Quantitative results on 8$\times$ super-resolution on the bicycle and stump scenes from MipNeRF-360 dataset.}
    \label{tab:x8}
\end{table}

Table \ref{tab:ablation} summarizes the quantitative results. Each component yields incremental improvements across all metrics, confirming the effectiveness of each design. Notably, the full internal-external joint guidance framework outperforms both the original and MV-Regulation-enhanced baselines, indicating that the joint guidance not only provides stronger multi-view consistency constraints but also enhances texture fidelity.
Figure \ref{fig:ablation} presents qualitative comparisons. The Mip-Splatting baseline suffers from blurred textures and missing details. MV-Regulation improves consistency but still lacks high-frequency recovery. Introducing external guidance sharpens textures but introduces local artifacts (e.g., in the grass regions). Adding internal guidance reduces these artifacts but retains some blurring. With mask-guided fusion, internal and external cues are effectively combined, producing sharp, artifact-free textures and accurate geometry, as observed in both rendered images and depth maps. Discrepancy mask analysis is provided in the Supplementary.

\subsubsection{Backbone Generalization Analysis}
To assess the generalizability of our joint internal-external guidance, we apply it to alternative backbone models. Specifically, for external guidance, SwinIR is replaced with PSRT \cite{rethinkingvsr}; for internal guidance, Mip-Splatting is replaced with Analytic-Splatting \cite{analytic}. We conduct experiments on two representative scenes, bicycle and stump, from the MipNeRF-360 dataset. As shown in Table \ref{tab:backbones}, our joint guidance consistently improves performance across different backbones on all metrics, confirming its strong robustness and backbone-agnostic generalization.

\subsubsection{Hyperparameter Sensitivity}
\begin{figure}
    \centering
    \includegraphics[width=0.7\linewidth]{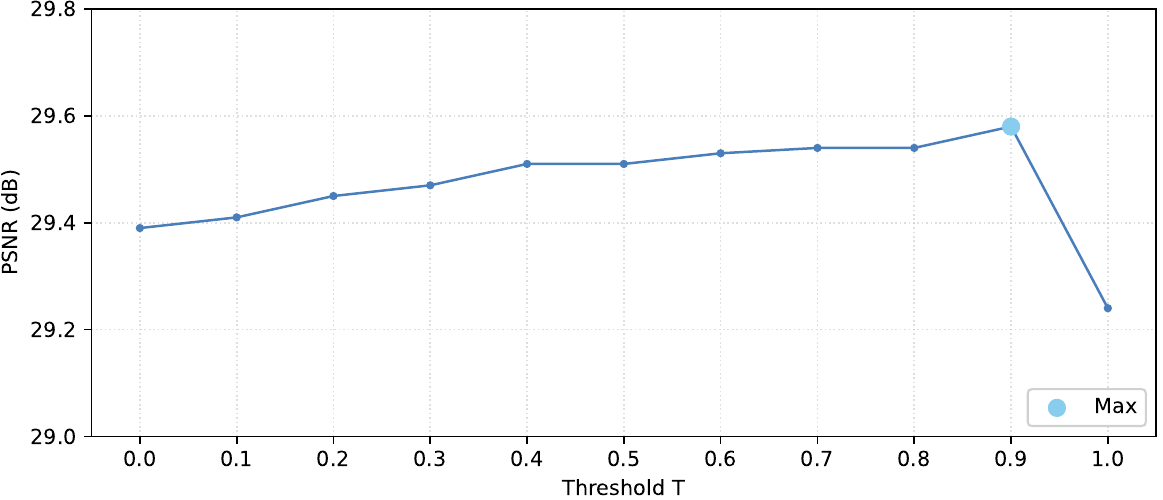}
    \caption{Threshold $T$ sensitivity in mask-guided texture integration on two scenes from the Mip-NeRF360 dataset.}
    \label{fig:threshold_sensitivity}
\end{figure}

To assess the effect of the threshold $T$ in mask-guided texture integration, we conduct a sensitivity analysis on two representative scenes from the Mip-NeRF360 dataset. The $T$ balances internal and external guidance: $T=0$ relies solely on internal supervision, while $T=1$ heavily trusts external guidance in ambiguous regions. As shown in Figure \ref{fig:threshold_sensitivity}, PSNR gradually improves as $T$ increases from 0 to 0.9, indicating that moderate external guidance enhances texture quality. The performance drop at $T=1$ suggests that excessive reliance on external priors introduces artifacts. These results confirm that IE-SRGS is robust to threshold selection over a wide range and further validate the effectiveness of integrating internal knowledge.

\subsubsection{Efficiency Analysis}
To assess the efficiency of IE-SRGS, we analyze the runtime of each component and compare it with the SOTA method SRGS \cite{feng2024srgs}. As shown in Table \ref{tab:runtime}, although IE-SRGS includes additional modules such as internal training and depth estimation, the total training time increases only slightly, by 7-8 minutes, demonstrating the efficiency of our design. Importantly, IE-SRGS achieves significantly faster inference, with 260 FPS on NeRF Synthetic and 119 FPS on MipNeRF-360, compared to 191 FPS and 92 FPS for SRGS. This gain is due to faster convergence enabled by our joint internal-external guidance and mask-guided integration. These results show that IE-SRGS achieves high-quality reconstruction with minimal added cost and superior runtime efficiency.

\subsubsection{Scaling Robustness Analysis}
To assess the scalability of IE-SRGS, we perform an additional 8$\times$ SR experiment on the MipNeRF-360 dataset, focusing on two representative scenes: bicycle and stump. Existing SOTA methods rarely report results under such extreme scaling, our experiment offers a more rigorous evaluation of model robustness. Table \ref{tab:x8} presents the quantitative results. IE-SRGS consistently outperforms both the baseline and SOTA method across all metrics, without requiring scene-specific fine-tuning. These results further highlight the strength of our internal–external guidance framework in preserving reconstruction quality even under large magnification factors.

\section{CONCLUSION}
We proposed IE-SRGS, a novel framework for 3D SR that integrates external and internal knowledge to optimize 3DGS. By combining HR detail priors with cross-view consistency and scene adaptation, IE-SRGS achieves high-fidelity 3D reconstruction from LR inputs. Extensive experiments demonstrate that IE-SRGS consistently outperforms SOTA methods and closely approaches the performance of HR upper bounds. This work lays the foundation for future research on unified frameworks for 3D low-level tasks and more effective internal-external knowledge integration.



\bibliography{aaai2026}

\newpage
\appendix

\section*{Appendix}\label{sec:reference_examples}

\section{Comparison of Rendered Depth Maps}
\begin{figure}[ht!]
\centering
\includegraphics[width=\columnwidth]{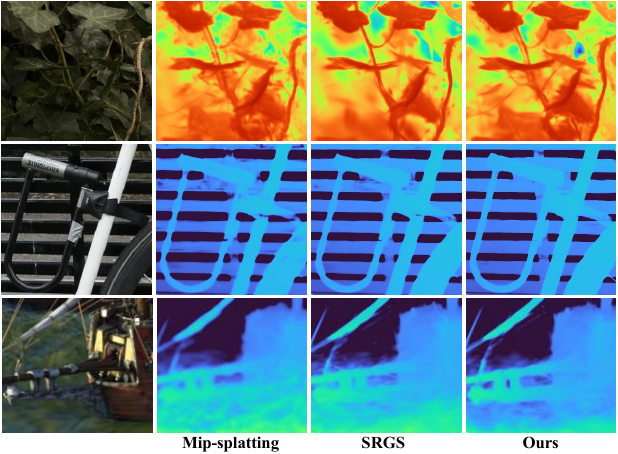}
\caption{Qualitative comparison of rendered depth maps by different methods for 4× 3D super-resolution on both synthetic and real-world datasets.}
\label{fig:depth}
\end{figure}

To address the lack of geometric detail in low-resolution 3D Gaussian Splatting (3DGS), we introduce, for the first time, a geometry optimization branch into the 3DGS super-resolution. This branch enforces explicit geometric constraints, guiding the reconstruction toward structurally consistent high-resolution scenes. Figure \ref{fig:depth} shows qualitative comparisons of rendered depth maps. Our method, IE-SRGS, produces notably cleaner and more coherent depth structures than state-of-the-art methods. These results highlight the effectiveness of geometry-aware optimization in enhancing structural fidelity and preserving fine-grained spatial details in 3DGS super-resolution.

\section{Robustness to Depth Estimator} 
To assess the robustness of our method to the choice of depth estimation model, we conducted an ablation study on two Mip-NeRF360 scenes using three representative depth estimators: DepthAnythingV2, DepthAnythingV2-Small, and DepthPro. As shown in Table \ref{tab:depth}, our method achieves consistently high PSNR, SSIM, and low LPIPS across all variants, with only marginal performance differences. These results demonstrate that our framework does not rely on a specific depth estimator and maintains strong performance across a range of depth qualities and runtime profiles.

\begin{table}[h]
    \centering
    \small
    \setlength{\tabcolsep}{3pt}
    \begin{tabular}{lcccc}
    \toprule[0.8pt]
    Depth Estimators & PSNR$\uparrow$ & SSIM$\uparrow$ & LPIPS$\downarrow$ & Runtime \\
    \midrule[0.8pt]
    DepthAnythingV2 & 29.58 & 0.889 & 0.205 &48.3s\\
    DepthAnythingV2 SMALL & 29.53 &0.887 &0.206 &29.3s \\
    DepthPro & 29.61  &0.889 &0.203 &80.1s\\
    \bottomrule[0.8pt]
    \end{tabular}
    \caption{Quantitative comparison of our method using different depth estimators on Mip-NeRF360 scenes.}
    \label{tab:depth}
\end{table}

\section{Effectiveness of Mask-Guided Fusion}
\begin{figure}[ht!]
\centering
\includegraphics[width=\columnwidth]{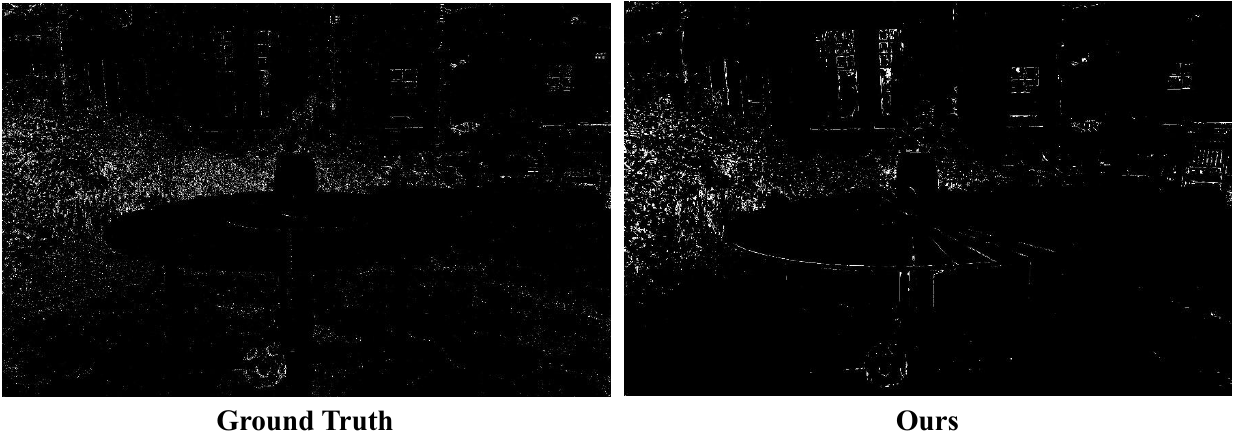} 
\caption{Comparison of discrepancy masks generated from ground-truth high-resolution image and external guidance (Ground Truth) versus from internal-external guidance (Ours).}
\label{fig:mask}
\end{figure}

We introduce a mask-guided fusion mechanism to adaptively integrate internal cues and external priors for 3DGS super-resolution. This design enables spatially selective guidance, resulting in sharper textures, more accurate geometry, and reduced artifacts, as demonstrated in the rendered RGB images and depth maps (Figures 3, 4, and 5 in the main paper). Figure \ref{fig:mask} compares our predicted discrepancy mask with the ground-truth mask derived from the high-resolution image and external guidance. The strong alignment between them confirms the reliability of our mask estimation strategy. These results demonstrate the effectiveness of mask-guided fusion in enhancing spatial adaptivity and structural fidelity, validating its critical role within the IE-SRGS framework.

\end{document}